%% file: sample-authordraft.tex
\documentclass[eat,twocolumn]{jmlr}

\usepackage{longtable}
\usepackage{booktabs}
\usepackage[load-configurations=version-1]{siunitx}
\usepackage{multirow}
\usepackage{soul,color,xcolor}
\usepackage{algorithm}
\usepackage[algo2e,ruled]{algorithm2e}
\usepackage{color}

\jmlrvolume{}
\firstpageno{1}

\jmlryear{2022}
\jmlrworkshop{Machine Learning for Health (ML4H) 2022}

\AtBeginDocument{%
  \providecommand\BibTeX{{%
    \normalfont B\kern-0.5em{\scshape i\kern-0.25em b}\kern-0.8em\TeX}}}

\title{Looking for Out-of-Distribution Environments in Multi-center Critical Care Data}

  \author{\Name{Dimitris Spathis\nametag{\thanks{Work done while at Microsoft Research Cambridge. DS is now with Nokia Bell Labs Cambridge and is affiliated with the University of Cambridge}}} \Email{ds806@cl.cam.ac.uk}\\
  \addr University of Cambridge, UK
  \AND
  \Name{Stephanie L. Hyland} \Email{stephanie.hyland@microsoft.com}\\
  \addr Microsoft Research
Cambridge, UK
 }

\begin{document}
\maketitle
\begin{abstract}

  Clinical machine learning models show a significant performance drop when tested in settings not seen during training. Domain generalisation models promise to alleviate this problem, however, there is still scepticism about whether they improve over traditional training. In this work, we take a principled approach to identifying Out of Distribution (OoD) environments, motivated by the problem of cross-hospital generalization in critical care. We propose model-based and heuristic approaches to identify OoD environments and systematically compare models with different levels of held-out information. We find that access to OoD data does \textit{not} translate to increased performance, pointing to inherent limitations in defining potential OoD environments potentially due to data harmonisation and sampling. Echoing similar results with other popular clinical benchmarks in the literature, new approaches are required to evaluate robust models on health records.

\end{abstract}



\section{Introduction} 

Machine learning models have achieved remarkable results in critical care \citep{hyland2020early} but most models are evaluated on data similar to what they have been trained with and deployed models in healthcare tend to perform worse when compared to the training phase \citep{castro2020causality, johnson2018generalizability}. This can be due to \textit{temporal shifts} \citep{guo2022evaluation, nestor2019feature} or new types of data \citep{finlayson2021clinician}. Recent domain generalization methods attempt to address these issues \citep{arjovsky2019invariant, li2018learning}, but they make strong assumptions about the nature of the dataset shifts. Previous works on critical care data concluded that the selected environments were \textit{not} OoD and instead resorted to inducing artificial shifts \citep{zhang2021empirical}. It is not clear however, whether these particular environments reflect real distribution shifts to generalize in different hospital settings. 

In this work we take a principled approach with regards to identifying OoD environments in real-world critical care data by proposing a model-based leave-one-hospital-out method and cross-sectional feature splits. We identify a set of potential OoD hospitals and analyze their characteristics in various dimensions including region, size, demographics, and teaching status, findings which could help future works using the same datasets. Then, we conduct extensive experiments with models on three different levels of access to OoD environments and assess the impact of data subsampling and model size in the task of mortality prediction. Despite the varying levels of performance drop in the first step of OoD candidate generation, we show that there is no significant performance improvement when using data from the OoD environment(s), which motivates further research on the suitability of this benchmark dataset for evaluating robust clinical models.

\section{Methods}

Following the notation introduced by Zhang et al. \citep{zhang2021empirical}, we assume labelled data $ \{(x_i^e,y_i^e )\}_{i=1}^n $, with $n$ examples in total, sourced from distinct training environments $e$. The goal of a robust predictor $f$ is to minimize the risk $R^e(f)$ (or loss) across all possible environments. This is usually assessed by reporting a performance metric in unseen test environment(s) $R^{e_{test}}(f)$ and associated validation environments $R^{e_{val}}(f)$. 

\subsection{Training scenarios} \label{training_scenarios} We compare the performance of three models with different levels of access to unseen OoD environments. The first one is traditional in-domain training and we report the performance of two "\textit{oracles}" which have access to the test environment(s) during
training:

\noindent \textbf{ERM} (Empirical Risk Minimization): traditional training where data from all training environments are pooled together.
 
\noindent  \textbf{ERMID}: an ERM model trained on the training split of the test environment(s). Assuming sufficient data from the test environment, we would expect this model to perform well, as it does not suffer from distribution shift.
 
\noindent  \textbf{ERMMerged}: an ERM model trained on the combination of all environments: training \textit{and} test environment(s). This model can be seen as an upper bound of performance when one has access to all data, which is an unrealistic scenario.

The difference in performance between the oracles and the ERM model is a proxy measure of how distinct the new environment is from the training environments, or in other words, a measure of OoD-ness of the environment.

\subsection{Environment splits}
\label{envsplits}

We consider the following environment splits to evaluate the impact of the model selection to the generalization performance. We designate disjoint training, validation, and test splits, and each environment is assigned to one split (also see Figure \ref{fig:splits}). Within each split, there are three train/validation/test sets. In other words, every environment (hospital)\footnote{An environment doesn't \emph{always} align with hospital, however in most cases in this work --except from Appendix Section \ref{envs_demographics}--, an environment is a hospital.} belongs to only one split and is further split to three sets. The data from the validation environment is used only for early stopping and hyperparameter tuning as needed. 
 
Each model employs the following configuration:

\noindent  \textbf{Training only}: \textit{ERM} uses the pooled training splits of the training environments, the validation split of the validation environment(s) for model selection, and the testing split of the test environment(s) for OoD generalization.

\noindent  \textbf{Testing only}: \textit{ERMID} uses the training split of the test environment, the validation split of the test environment for model selection, and the testing split of the test environment for OoD generalization.

\noindent  \textbf{All environments}: \textit{ERMMerged} uses the pooled training splits of the combined training, validation and test environments (all), the combined validation splits of all environments, and the testing split of the test environment(s) for OoD generalization.

Note that the test split of the test environment is held out from all models, and can be used to compare performance.

\begin{algorithm2e}
\SetAlgoLined
  \SetKwInOut{Input}{Input}
   \SetKwInOut{Output}{Output}
 \Input{$ \{(x_i^e,y_i^e )\}_{i=1}^n $ (input data),  $\{e_m\}_{m=1}^{|E|}$ (environments), $T$ (threshold), $P$ (env performance measure)}
 \Output{ $\{f^{E\setminus m}\}_{m=1}^{|E|}$ (predictors), $e_m' \subseteq e_m$ (candidate envs.)	 }
 \For{$e_m$ in $\{e_1, \dots, e_{|E|}\}$}{
  train predictor $f^{E\setminus m}$ on environment $E \setminus m$ (with $ \{(x_i^{e},y_i^{e} )\}_{i=1; e \neq e_m}^{n} $)\;
  evaluate predictor based on performance measure $P$, to obtain $P^m_{\text{in-domain}}$ (on environment $E \setminus m$), and $P^m_{\text{out-domain}}$ (on environment $m$)\;
  compute $P_{rank}^m = P^m_{in-domain} - P^m_{out-domain}$\;
 }
 rank predictors $f^{E \setminus m}$ by $P_{rank}^m$\;
 apply threshold $T$ to $P_{rank}$  and generate candidate list $e_m'$ \;
 set $e_m'$ environments as test set for OoD validation\;

 \caption{Model-based OoD identification}
 \label{algorithm}

\end{algorithm2e}

\subsection{Model-based OoD environment identification}

Motivated by the problem of cross-hospital transfer of clinical models, we design a model-based OoD environment identification approach. To this end, we use model performance to evaluate OoDness. This is based on the observation that the model can be a good reflection of the underlying training data it has been trained on, including the non-linear interactions of multiple features that could be impossible to capture with bespoke exclusion criteria. 
Considering that it is impossible to know beforehand which environment is going to be OoD, we propose a Leave-One-Hospital-Out (LOHO) training setup to conduct an exhaustive search over possible environments. This way, we move away from single exclusion criteria (e.g. patients with high blood pressure), and instead we use the model performance as proxy for OoDness. To facilitate this, we are considering unique hospitals as candidates for those environments. 

Specifically, we iterate through all $m$ hospitals, treating each one as the test environment in turn. We use the remaining $m-1$ hospitals as the train environment, training $m$ predictors in a leave-one-out fashion. For each trained model, we compute the Out-of-Domain test performance ($P_{\text{out-domain}}$); computed on the test set of the test split (unseen data from the unseen environment) as well as the In-domain test performance ($P_{\text{in-domain}}$);, computed on the test set of the train split (unseen data from the same environment(s)).

We consider the difference between these as indicative of the `OoD-ness' of the left-out hospital (test environment). We can then rank each hospital according to this difference:

\begin{equation}
    P_{rank}^{m} = P_{in-domain} - P_{out-domain}
\end{equation}

A threshold $T$ is applied to $P$ as a cutoff value to select candidate environments. In practical terms, $T$ can be a quantile cutoff of the environments' distribution and its value can be traded off with the size of the resulting test set. We should take into account that in most ML experiments, the test set corresponds commonly to 20\% of the available data. Algorithm~\ref{algorithm} describes the steps of the model-based OoD environment detection approach.

The test and validation environments come from the thresholding step of Algorithm \ref{algorithm}. The intuition behind this choice is that validation environments should show OoD qualities, similar to those on the test set. We discuss more about these parameters in the next sections.

\subsection{Comparing models on equal terms} Considering the different environment splitting strategies we discussed in \ref{envsplits}, we acknowledge that ERMID is typically trained on significantly less data compared to ERM and ERMMerged. ERMID has only access to the candidate OoD environment, or in the case of critical care, the local hospital(s). Machine learning models tend to perform better when trained on more data, so it would be difficult to assess whether potential performance gains are due to the choice of the environment or merely because the model has seen more examples. To compare models on equal terms we must control for the training data set size. This would entail subsampling the datasets used with ERM and ERMMerged. A first approach would be to apply naive subsampling to match the size of ERMID with that of ERM \& ERMMerged. However, by doing so, there is the possibility of discarding data from entire environments. To mitigate that, we randomly select data within each environment (e.g. hospital) so that they all add up to match the size of ERMID.

\section{Data \& Experimental setup}
The eICU Database
includes data from patients in the Intensive Care Unit (ICU) spanning over 200,000 admissions from over 200 hospitals in the United States \citep{pollard2018eicu}. The ICU environment is information-rich and the literature has explored many different research questions \citep{sjoding2020racial, rocheteau2021temporal}. Given its large number of hospitals, we use it as our testbed for OoD environment identification.

\textbf{Mortality prediction}. In this work, we focus on mortality prediction, borrowing the experimental setup and cohort selection as defined in previous works \citep{zhang2021empirical, sheikhalishahi2020benchmarking}. This is a binary prediction task which aims to forecast ahead of time whether a patient will die while in hospital, using data from the first 48 hours of the stay. We include patients who are between 18-89 years old and who are still alive after the first 48 hours. For those with multiple ICU stays, only the first stay is included.

Each patient provides timeseries and static features. There are 10 continuous and 4 categorical timeseries features, as well as 3 numerical and 2 categorical static features. The observations are resampled to 1-hour windows and missing values are imputed from the previous observation with forward filling. The resulting dataset has 30,691 patients from 208 hospitals. To ensure there are enough datapoints per environment, we further exclude hospitals with $<$50 patient stays, ending up with 29,082 patients from 127 hospitals. Each hospital belongs to one of the following US regions: Midwest, West, Northeast,  South, and the placeholder Missing region for no data.

\textbf{Model architecture}. To allow for comparisons with previous work \citep{zhang2021empirical}, we employ the same model architecture which consists of bidirectional Gated Recurrent Unit \citep{chung2014empirical} layers followed by a linear layer for classification with a 2-unit output. The categorical variables
pass through individual embedding layers per variable, and standard scaling is applied to continuous features. The GRU layers receive the timeseries features along with appended static features at each timestep. The model hyperparameters for our experiments are provided in Appendix \ref{appendix}.

\textbf{Assigning OoD candidates to splits}. After applying Algorithm \ref{algorithm} to the data, we end up with a candidate list which contains the hospital IDs sorted by the lowest $P_{rank}$. To assess whether these hospitals are actually OoD we conduct experiments using the training scenarios as described in \ref{training_scenarios}. To this end, we assign hospitals to train/validation/test sets. We argue that the worse-performing hospitals in terms of $P_{rank}$ should go to the test and validation set. In practice, we use the lowest 20\% quantile of the $P_{rank}$ distribution to generate the OoD candidate list. Out of those hospitals, we assign the largest ones to the test set and the rest to the validation set, approximating a final split ratio of 85\%/10\%/5\% across the three sets, which is common in ML experiments. The full environment list is available in the Appendix \ref{appendix}.

\textbf{Evaluation}. Given the binary prediction task, we employ the Area Under the Receiver Operating Characteristic curve (AUC-ROC), which can handle class imbalances. We note that the original dataset shows a 11\% mortality rate (positive class) \citep{zhang2021empirical}. We train for 100 epochs or until AUC stops improving for 7 consecutive epochs on the validation set. Unlike previous work, we employ bootstrapping by sampling  with replacement on the test set (500 repetitions) to calculate 95\% confidence intervals.

\subsection{Results}

\textbf{Are these environments really OoD?} After assigning the candidate hospitals to the three splits, we train the ERM, ERMMerged, and ERMID models to investigate whether access to test environments impacts OoD generalization. The statistics of the sets are provided in Table \ref{tab:sets_statistics}.

Since the ERM and ERMmerged models have access to more environments, they are trained on more data which could potentially impact the performance and comparisons. On that account, we evaluate an imbalanced and a resampled variant. Despite the test environment being putatively OoD by the LoHo approach, both subsampling variants show no \emph{significant} difference across the three models (Table \ref{tab:resampling}). Predictably, the resampled experiment yields lower AUCs given that all models have access to an order of magnitude fewer datapoints (data sizes available in the Appendix \ref{appendix}).

Focusing on the resampled variant which we consider to be fairer compared to other setups, we also investigate the impact of model overparameterization. We evaluate two model variants by scaling down the model size (reducing the GRU layers and units). More details about the networks are available in the Appendix \ref{appendix}. In Table \ref{tab:param}, we see that reducing the network size leads to improved AUC, slightly outperforming the Imbalanced variant of Table \ref{tab:resampling}. In other words, reducing the model complexity trades off the smaller training size in ERM and ERMMerged.

\input{table_resampling}

Considering the broad confidence intervals derived from bootstrapping in the results described above, we also tested computing variance by changing the random seed of training/model initialisation, reporting mean and standard deviation. This experiment yielded an AUC of 0.62 ($\pm$0.01) for ERM, 0.61 ($\pm$0.04) for ERMMerged, and 0.58 ($\pm$0.05) for ERMID. In all cases,  ERMID underperforms the rest models, hinting that these test environments are not OoD. We note that models restricted to local-hospital information were found to underperform in previous works with private datasets and very low number of hospitals, however they were not deliberately selected as potential OoD environments  \citep{wiens2014study}.

\input{table_parameterization}

\section{Discussion}

We believe that the results presented above motivate further work on stress-testing OoD environments and models in clinical settings, as well as call for theoretical advances in automating the discovery of such environments. We want to stress that the purpose of this work is to highlight the inherent limitations in popular clinical datasets frequently used as benchmarks, rather than to propose new ML models.

Our study has the following limitations. Even though bootstrapping with replacement is supposed to be the golden standard in reporting medical results and comparing across different statistical models, the small test environment dataset sizes result in large confidence intervals and inconclusive findings. However, since real-world applications of domain-invariant methods are liable to be deployed in scenarios with limited test environment data, methods for quantifying dataset shift (e.g. OoDness) even on limited data are necessary. An alternative comparison would consider paired bootstrap confidence intervals for a pair of models on each bootstrap sample.

Another limitation is that the prediction task itself might contribute to the problem of identifying OoD environments in the eICU. The conditional probability of mortality given readily-measured variables in an ICU may not genuinely shift significantly between hospitals. For future work, we plan to apply our model-based approach to tasks more likely to be susceptible to operational factors, such as the prediction of Length of Stay in the ICU. Combined with the Mortality task, we expect to identify environments and hospitals with more 'distinct' characteristics that differ from the 'average' hospital. 


\section{Conclusion}
Here we proposed a framework for OoD detection in multi-center critical care data. We argued that we lack principled ways to identify "natural" OoD environments and conducted extensive experiments in a Leave-One-Hospital-Out fashion by benchmarking three models with different levels of test-set access. We found that access to OoD data does not improve test performance, which points to inherent limitations in defining OoD environments in the eICU Database, potentially due to extensive data harmonization and processing applied during its collection. Our alternative training scenarios employed cross-sectional features as potential OoD environments and hinted that this approach might be promising in specific feature combinations that however require domain knowledge. All in all, echoing similar results with other established clinical benchmarks in the literature, we believe that new methodological approaches along with new benchmarks are required for the evaluation of robust ML models in critical care.

\bibliography{sample-base}


\appendix
\onecolumn

\renewcommand\thefigure{Suppl. \arabic{figure}}    
\renewcommand\thetable{Suppl. \arabic{figure}}    
\setcounter{figure}{0}    
\setcounter{table}{0}

\section{Supplementary material}
\label{appendix}

\subsection{Hospital environments} 
Environment splits as a result of applying the LOHO procedure and the threshold of Algorithm \ref{algorithm}. The numbers correspond to hospital IDs in the eICU.

\begin{itemize}

 \item \textbf{Training environments}: \texttt{[217, 272, 224, 253, 434, 157, 199, 220, 403, 252, 122, 95, 440, 207, 167, 148, 392, 281, 458, 365, 181, 226, 382, 271, 204, 141, 436, 244, 279, 407, 413, 205, 188, 383, 110, 420, 439, 337, 248, 429, 183, 140, 423, 443, 421, 227, 428, 144, 277, 318, 245, 389, 445, 208, 310, 165, 268, 264, 146, 200, 419, 411, 388, 202, 424, 180, 307, 435, 259, 197, 63, 79, 142, 449, 206, 249, 394, 152, 331, 171, 358, 400, 422, 444, 243, 416, 312, 176, 254, 194, 143, 338, 396, 73, 301, 300, 269, 280, 336, 402, 433, 210, 66, 215]}.

\item \textbf{Validation environments}: \texttt{[387, 390, 397, 417, 452, 459, 71, 92]}.

\item \textbf{Test environments}: \texttt{[108, 154, 184, 195, 196, 198, 256, 275, 283, 345, 384, 386]}.

\end{itemize}

\noindent Three environments that had low patient stays were excluded from the splits because we could not calculate evaluation metrics: \texttt{[353, 281, 391]}.

\subsection{Resampling experiment}

We compare two different configurations of training, one with more data on the ERM and ERMMerged environments, and one with environment-based resampled/matched data:

\begin{itemize}
\item \textbf{Original trainset}: \texttt{ERM=17962, ERMMerged=20337, ERMID=1525}. 

\item \textbf{Resampled trainset}: \texttt{ERM=1554, ERMMerged=1524, ERMID=1525}.

\end{itemize}

\noindent Small differences in the resampled sets' sizes are due to the randomization process within environments which does not always match the ERMID size.

\subsection{Feature names} The following features are used from eICU:
\begin{itemize}
\item \textbf{Temporal}: \texttt{Heart Rate, MAP (mmHg), Invasive BP Diastolic, Invasive BP Systolic, O2 Saturation, Respiratory Rate, Temperature (C), glucose, FiO2, pH, GCS Total, Eyes, Motor, Verbal}.

\item \textbf{Static}: \texttt{Admission height, Admission weight, age, Apache admission dx, gender}.
\end{itemize}

\subsection{Model hyperparameters}
\begin{itemize}

\item \textbf{Large model}: \texttt{batch size = 16, \#parameters: 783,538, GRU layers = 3, GRU hidden dimension = 128, embedding dimension = 16, learning rate = 0.001.
}

\item \textbf{Small model}: \texttt{batch size = 8, \#parameters: 34,162, GRU layers = 1, GRU hidden dimension = 32, embedding dimension = 16, learning rate = 0.001.
}

\end{itemize}

\section{Motivation} 

Machine learning models have achieved remarkable results in critical care \citep{hyland2020early} but most models are evaluated on data similar to what they have been trained with. Common evaluation practices involve splitting the training and testing sets based on some criteria, ranging from completely random sampling to group-based cross-validation, especially when working with human-centered applications. Still, deployed models in healthcare tend to perform worse when compared to the training phase, due to the dissimilarity between
training (in-distribution) data and data that
the model is applied to after deployment \citep{castro2020causality, johnson2018generalizability}. 

Multiple factors can influence the deployment performance of a clinical model. Observing differences on withheld data from previous or later years constitutes a \textit{temporal shift} and has found to be an important factor of performance drop \citep{guo2022evaluation, nestor2019feature}. This can be due to various reasons: new types of data collection devices or terminologies (ICD-9 vs ICD-10 codes), internal changes in variable definitions when adopting a new Electronic Health Records (EHR) platform, changes in disease incidence (e.g., COVID-19 prevalence), or changing demographics (e.g., through hospital mergers) \citep{finlayson2021clinician}.

Machine learning models are known to struggle to generalize OoD, failing in unexpected ways
when tested outside of the training samples domain. For instance, self-driving cars are affected by variations in light or weather \citep{dai2018dark}. This undesirable behaviour can be attributed to inherent limitations of neural networks trained on pooled data that learn easier-to-fit spurious correlations instead of the causal factors of
variation in data. A notable example are vision models focusing on the background of the image instead of the respective object (cows in `common' contexts such as alpine pastures are detected correctly, while cows in uncommon contexts such as the beach are not) \citep{beery2018recognition}. Therefore, the unpredictable behaviour of machine learning models given OoD inputs
 constitutes a significant obstacle to their deployment in critical applications such as in healthcare.
 
 Recent methods attempt to address these issues. In particular, the area of `domain generalization' assumes access to multiple `environments' during training, each of them containing samples in different conditions. A successful model should learn the invariances across these environments that will then generalize to held-out test domains. However, despite the introduction of multiple domain generalization methods \citep{arjovsky2019invariant, li2018learning}, there is evidence that they do not outperform traditional domain-unaware training \citep{gulrajani2020search}. 

A key challenge in this area is that domain generalization methods typically make strong assumptions about the nature of the dataset shift. When these assumptions are violated, it is not surprising that methods may not demonstrate improved performance. However, it is not always clear whether such assumptions can be applied to \emph{real-world} dataset shifts. One of the first works to apply domain generalization methods to critical care, concluded that the selected environment was \textit{not} OoD and instead resorted to inducing artificial shifts such as adding noise or resampling \citep{zhang2021empirical}. It is not clear however, whether these particular environments reflect real distribution shifts and can generalize in different hospital settings.

\section{Further related work}

Despite the abundance of machine learning benchmarks for most data types, there are considerably fewer benchmarks to evaluate OoD detection models. Recent attempts to evaluate models on different environments include \textit{DomainBed} and \textit{WILDS} \citep{gulrajani2020search, koh2021wilds}, which have curated datasets ranging from textual data, to satellite and medical images. However, there is limited work on critical care and electronic health record benchmarks.

Due to this lack of standard OoD benchmarks, recent works introduce synthetic shifts to evaluate the generalization capabilities of models in unseen domains. The most relevant work to ours compared domain generalization models to traditional empirical risk minimization (ERM), using the eICU database \citep{zhang2021empirical}. One of the five available hospital regions ("South" region) was selected as OoD environment because its demographics appear to be the most distinct (mainly ethnicity). However, according to the authors, the performance of ERM on the eICU test set was on-par with the "oracles" that have access to this environment, indicating that this environment is likely not OoD. To overcome this limitation in this real-world dataset, the authors proposed a set of synthetic domain shifts such as noise-corrupted labels, feature-correlated corrupted labels, and biased subsampling. These artificial shifts were empirically found to be OoD, but the impact of domain generalization (DG) models was not significant over ERM. We posit that it is challenging to deconfound whether this is attributed to the models or the selected environments. Another important factor here is that even \emph{if} the DG methods had improved over ERM, because these were 'synthetic' OoD environments, it is not a given that those would work for real-world shifts. We should note that experiments with medical imaging data showed that OoD environments are easier to find but ERM still performs equally well. However, given our focus on critical care data, we believe that there is a research gap in appropriate benchmarks in this area and we build upon their framework by exploring more environments in a principled model-based approach.

A similar sentiment is echoed with the MIMIC database \citep{johnson2016mimic}, where studies on temporal shifts have been inconclusive. We note that MIMIC contains EHR data from 2008 to 2019. A common experimental setup is to train models on a snapshot of data and evaluate them on subsequent years. However, domain generalization models struggle to outperform domain-unaware models in EHR tasks including mortality, sepsis, invasive ventilation, and length of stay \citep{guo2022evaluation}. More importantly, this study compared models trained on each individual year to a single model trained on the first snapshot and tested on the subsequent years. The findings hint that temporal
dataset shift was not detected in three out of four tasks, where only the Sepsis task showed significant differences. This motivates the problem of a principled OoD identification method which is the main focus of our work.

Other works focus on coming up with OoD environments through exclusion criteria that are medically meaningful. This involves excluding groups based on demographics, splitting features related to a dynamic clinical status, or artificially creating OoD groups by withholding them during training \citep{zadorozhny2021out}. We extend this line of work by proposing combinatorial criteria (e.g. both gender and age). However, we attempt to identify natural OoD environments, whereas Zadorozhny and colleagues' scope didn't include comparisons to assess the OoD-ness of the environments. Its focus was on benchmarking density estimation models in the AmsterdamUMC Database, which includes a mortality task like in our case \citep{zadorozhny2021out}.

Another challenge in evaluating predictive models in clinical care is comparing across different medical centers \citep{desautels2017using}. As discussed above, most popular datasets such as MIMIC are single-center whereas the eICU Database spans more than 200 hospitals \citep{pollard2018eicu}. A systematic comparison study applied cross-validation across different hospitals in the eICU and assessed how well EHR models transfer to held out hospitals as compared to locally developed models \citep{johnson2018generalizability}. We incorporate a similar cross-validation scenario to identify OOD hospitals and assess transferability. Also, the recalibration process involved transferring the feature scalers (ranges of vitals etc.) to other hospitals, whereas more recent approaches applied pre-training and fine-tuning but there were no experiments across hospitals \citep{mcdermott2021comprehensive}. Last, a recent work focused on the generalization gap across hospitals in the eICU \citep{singh2022generalizability}, however the hospital-level analyses were restricted to the top 10 hospitals with the most stays and its overall goal was not to discover OoD environments.

\section{Further analysis}

\subsection{Ranking all environments} 
\begin{figure*}
    \centering
    \includegraphics[width=1\textwidth]{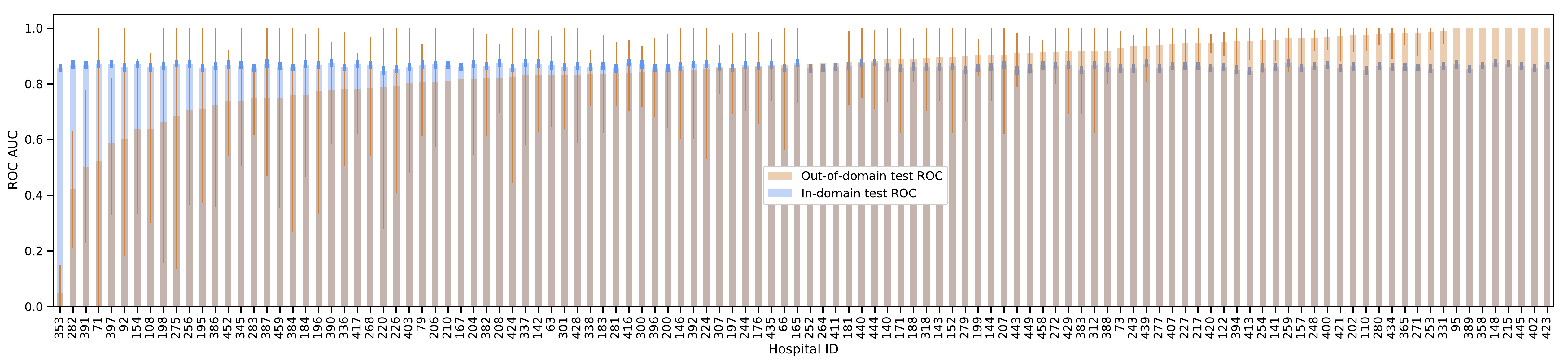}
    \vspace{-1cm}
    \caption{\textbf{Ranking all hospitals with Leave One Hospital Out (LOHO) training}. On the left tail we can see the miss-match between in-domain test ROC (blue) and out-of-domain test ROC (orange). These hospitals are OOD candidates because there is significant performance drop when tested in different domains. Darker blue and orange lines respectively denote 95\% bootstrap confidence intervals.  }
    \label{fig:LOOranking}
    \vspace{-0.7cm}
\end{figure*}

After applying Algorithm \ref{algorithm} to eICU according to the previous sections, we end up with a ranking of all hospitals sorted by those with the higher $P_{rank}$ to the lowest. In other words, these hospitals have a high In-domain but low Out-of-domain performance. In Figure \ref{fig:LOOranking}, we can see these hospitals on the left tail of the ranking along with their IDs. We note that there are hospitals on the right tail of the ranking where their Out-of-domain  exceeds the In-domain performance. These are mainly smaller hospitals with high mortality  imbalance which corrupts the evaluation metric. Similarly, some hospitals on the left tail exhibit out-of-domain AUCs near or below 0.5, reflecting metric instability. We also observe that the confidence intervals on the test set are significantly broader compared to the in-domain set which is more stable since it comes from the same environment(s) \footnote{As future work, one could investigate the extent to which these differences across hospitals are due to random variation by comparing the distribution to a simulated distribution under the null (i.e., all hospitals have the same expected performance).}. This is also because the in-domain test set is considerably bigger, so bootstrapping has lower variance.

\begin{figure*}
    \centering
    \includegraphics[width=0.5\columnwidth]{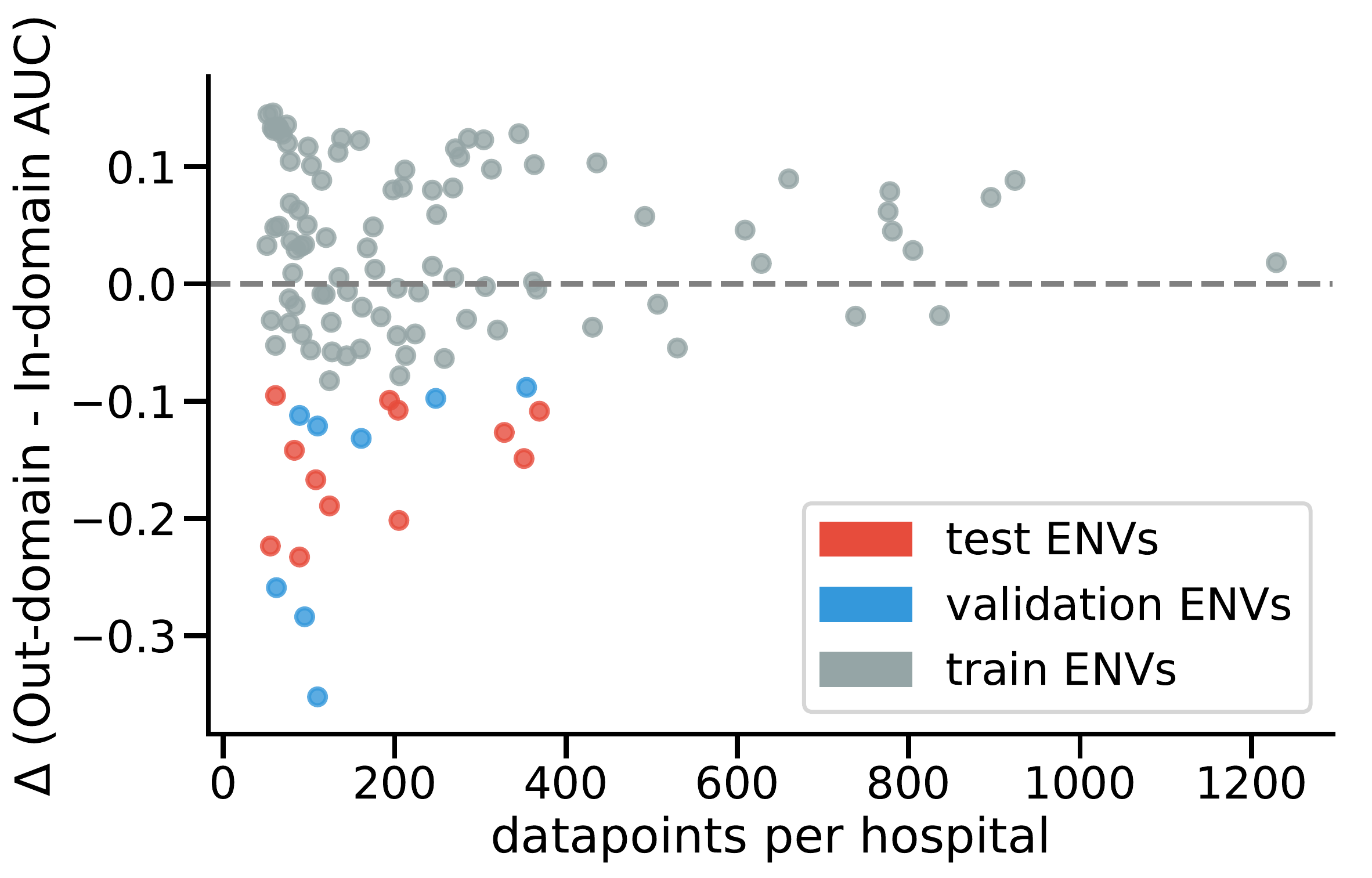}
    \caption{\textbf{Picking the test set}. Scatterplot of the assigned environment splits as a function of the performance delta between Out-of-domain and In-domain AUC on the LOHO experiment. The dashed line denotes the area where the two metrics are equal (In-domain=Out-of-domain). }
    \label{fig:splits_scatterplot}
    
\end{figure*}

In Figure \ref{fig:splits_scatterplot} we can see the overall performance $P_{rank}$ as a function of hospital size (datapoints per hospital), which shows that in the few large hospitals  there seem to be no OoD environment, whereas almost all potential OoD environments are concentrated on smaller hospitals featuring between 100 and 400 datapoints. 

\textit{A note on the LOHO experiment}: we acknowledge that different approaches could be explored. In particular, the idea of comparing ERM to ERMID for candidate generation could be promising, however, we may have to deal with overfitting issues due to different data sizes. In our current setup, after ranking all environments we train on source and evaluate on source vs. target. The source evaluation dataset is larger than the target one, which might explain the consistently high in-domain AUCs (blue bars in Figure \ref{fig:LOOranking}). The alternative approach would be to train two models: one trained on source vs. one trained on target; both evaluated on target. We note that this approach is identical to our training scenarios once we have identified the OOD environments and is the final step of our proposed method. Nevertheless, we understand the subtle difference that this approach could be also employed in the LOHO step. Our motivation for the in-domain vs out-domain ranking was that the OOD candidate generation training step should not have access to local/OOD hospital data (apart from testing). While we acknowledge the alternative idea as promising, it disregards this assumption.

\subsection{The characteristics of the worst-performing hospitals} 
\begin{figure}
\centering
\includegraphics[width=.35\textwidth]{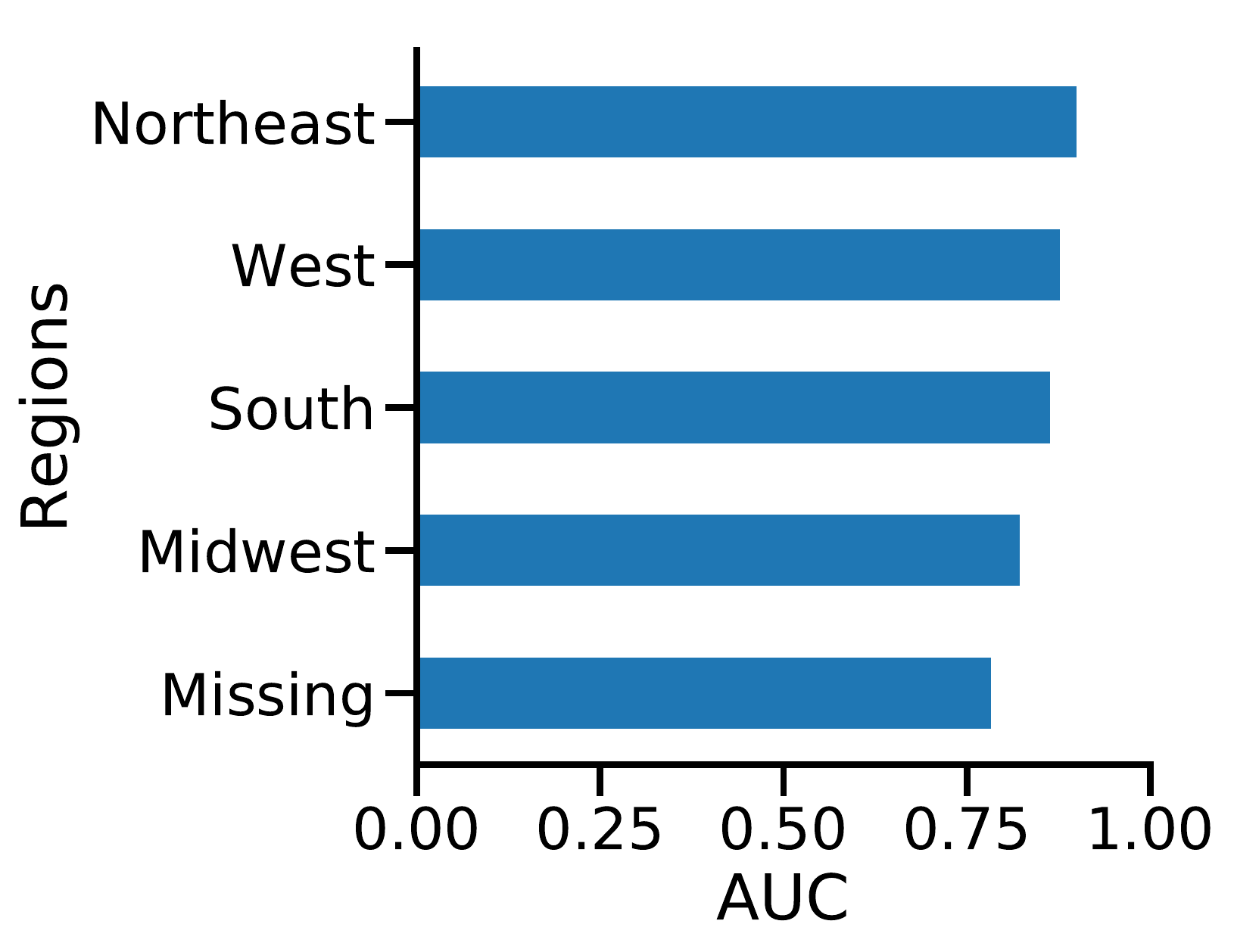}\hfill
\includegraphics[width=.28\textwidth]{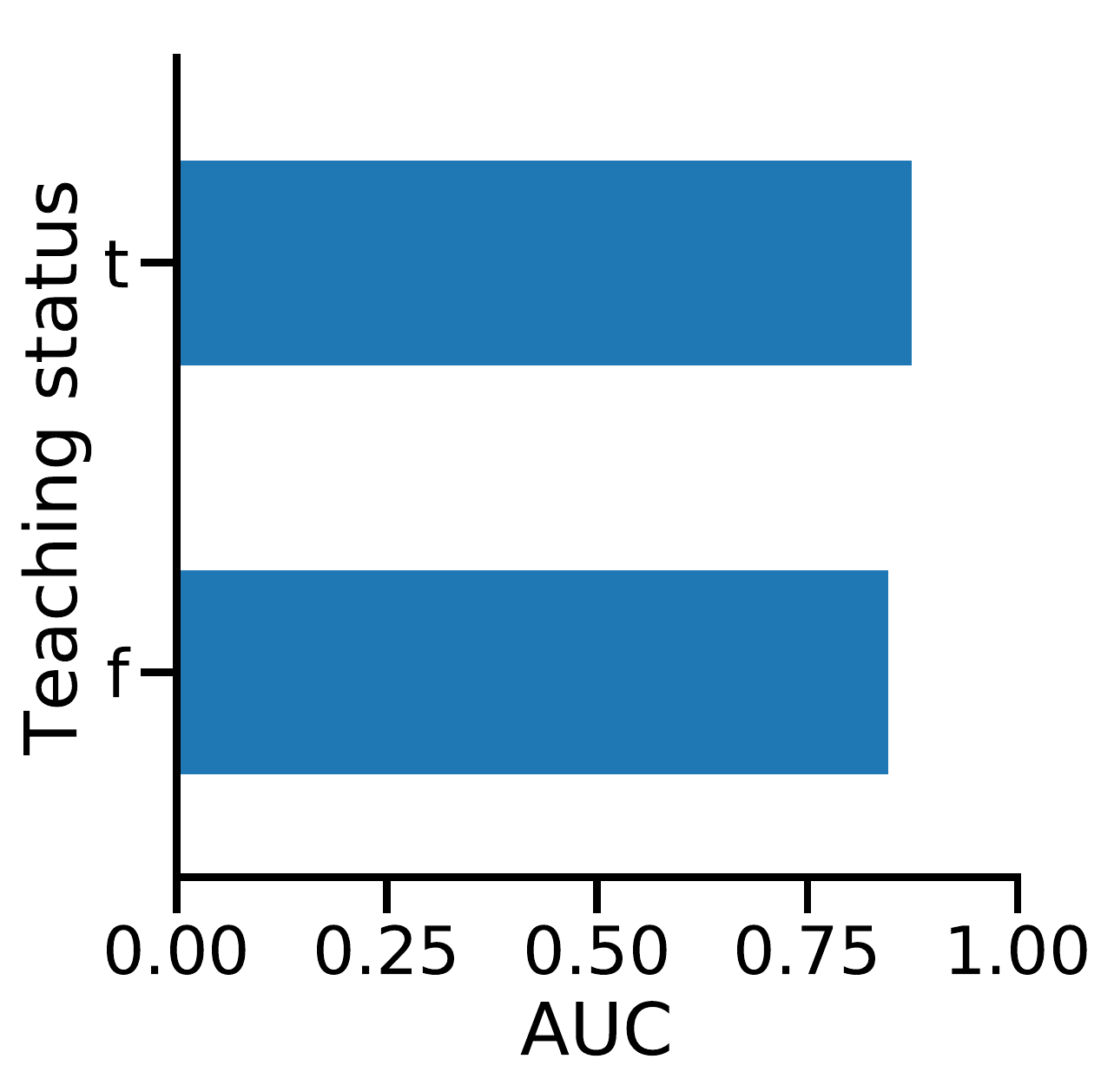}\hfill
\includegraphics[width=.35\textwidth]{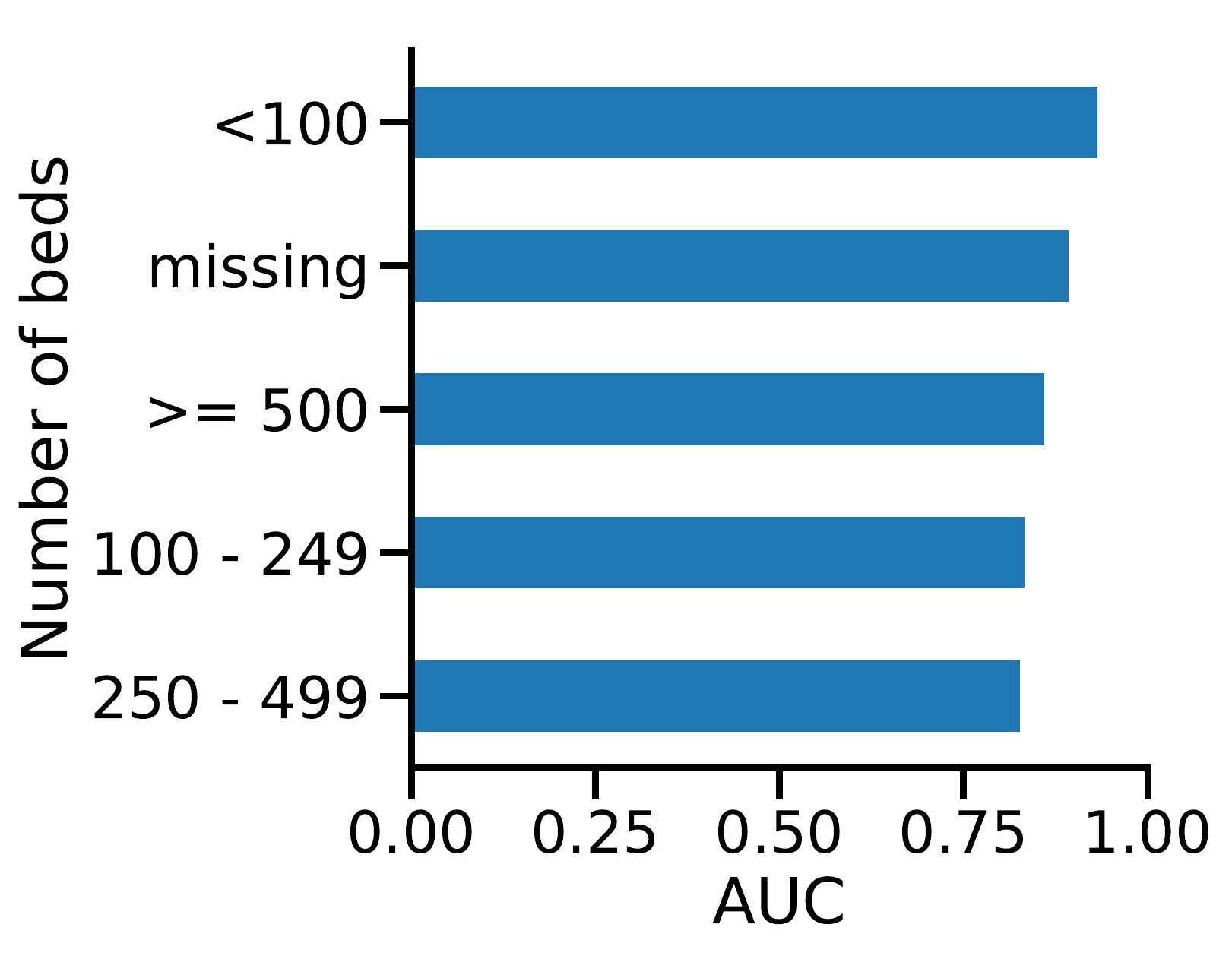}
\caption{\textbf{Grouped performance of hospitals after applying the LOHO procedure}. Instead of looking at individual OoD hospitals, here we examine whether the regions they come from, the teaching status, and the number of beds affect OoD performance.}
\label{fig:aggregate}

\end{figure}

In Figure \ref{fig:aggregate} we aggregate the LOHO results and group the hospitals by region, teaching status, and number of beds. The first observation is that there are notable differences between the regions. In particular, hospitals in regions 'Missing' and 'Midwest' tend to have lower test performance and therefore are potential OoD environments. On the other hand, the 'Northeast' region shows the highest performance, followed by 'West' and 'South'. This result also confirms the inconclusive findings of previous studies using the 'South' region as OoD \citep{zhang2021empirical}. By using our LOHO approach, \textit{we recommend to future studies employing region-level environments to focus on the 'Missing' and 'Midwest' environments} \footnote{Our focus is on hospital-level  rather than region-level environments and therefore we do not explore this direction.}. More interactions between the transferrability across the regions can be found in the literature \citep{singh2022generalizability}.

Beyond regions, we observe smaller differences between the teaching status of the hospital, with non-teaching ones performing worse. Last, we see that hospitals with 250-499 beds tend to perform worse, followed by those with 100-249 beds. It is noteworthy that the hospitals with $<$100 beds tend to perform better than every other category, pointing to non-linear correlations between the number of patient stays (or datapoints in Figure \ref{fig:splits_scatterplot}) and number of beds, in terms of OoD performance. This relationship should be investigated more in future works. However, another observation across these groupings is that the number of hospitals seems to explain some of these differences, for example the 'Northeast' region features only 13 hospitals as well as there are only  8 hospitals with $<$100 beds. Bigger groups (e.g. 250 - 499 beds) tend to include more hospitals which predictably introduces more noise and diverse patient demographics which impact test performance. 

\begin{figure*}[h]
\centering
\includegraphics[width=.75\textwidth]{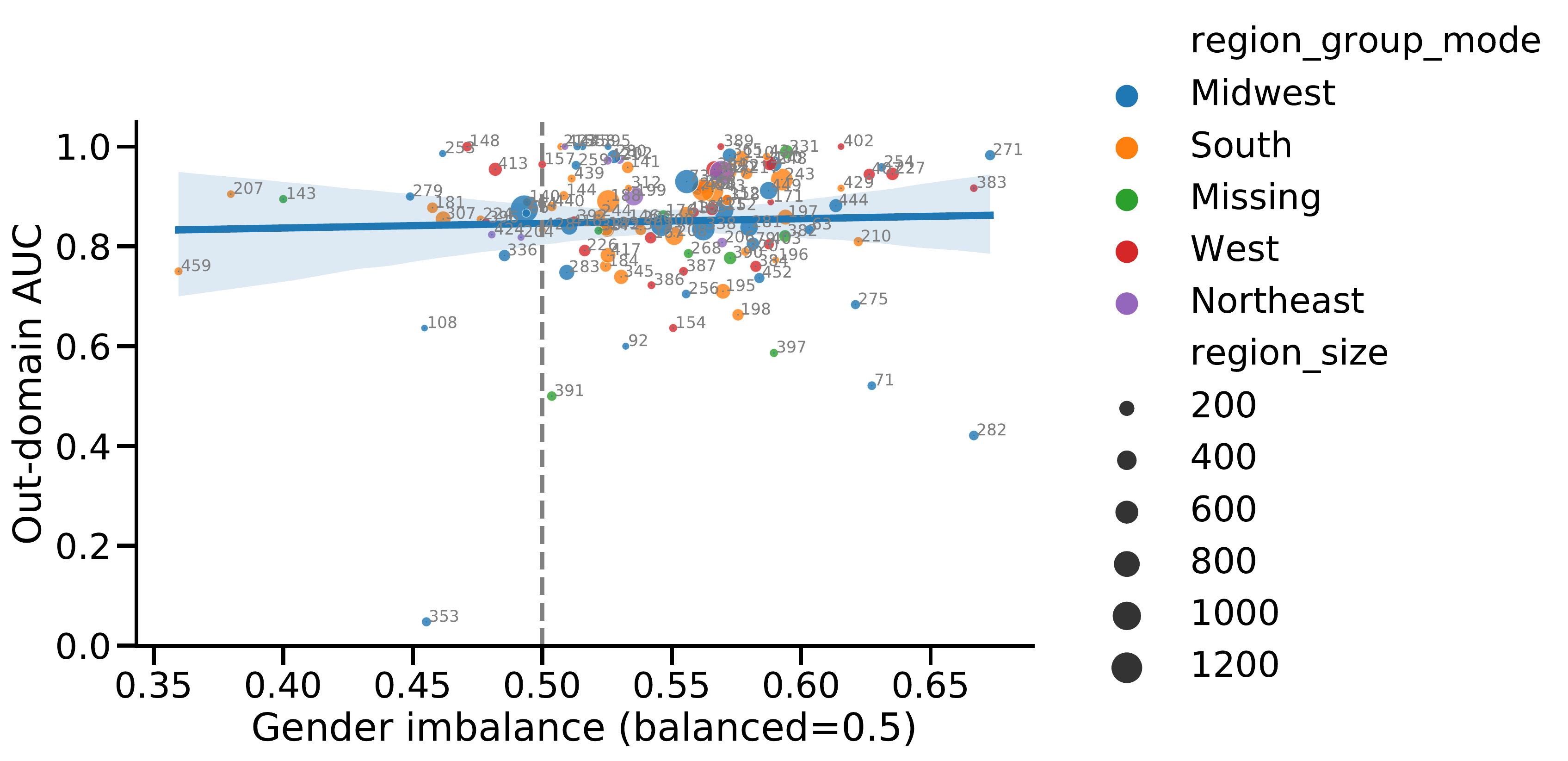}\hfill
\includegraphics[width=.48\textwidth]{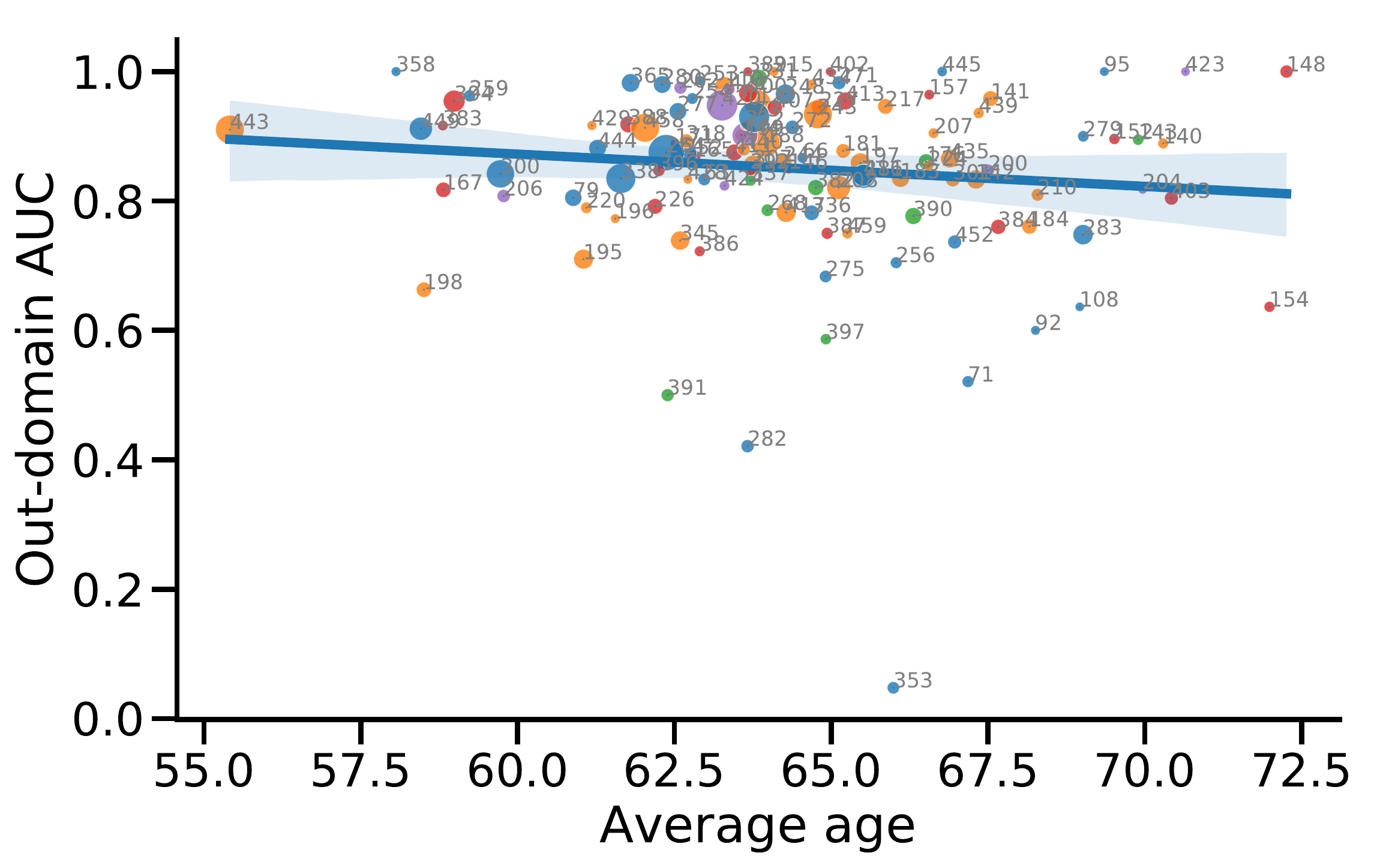}\hfill
\includegraphics[width=.52\textwidth]{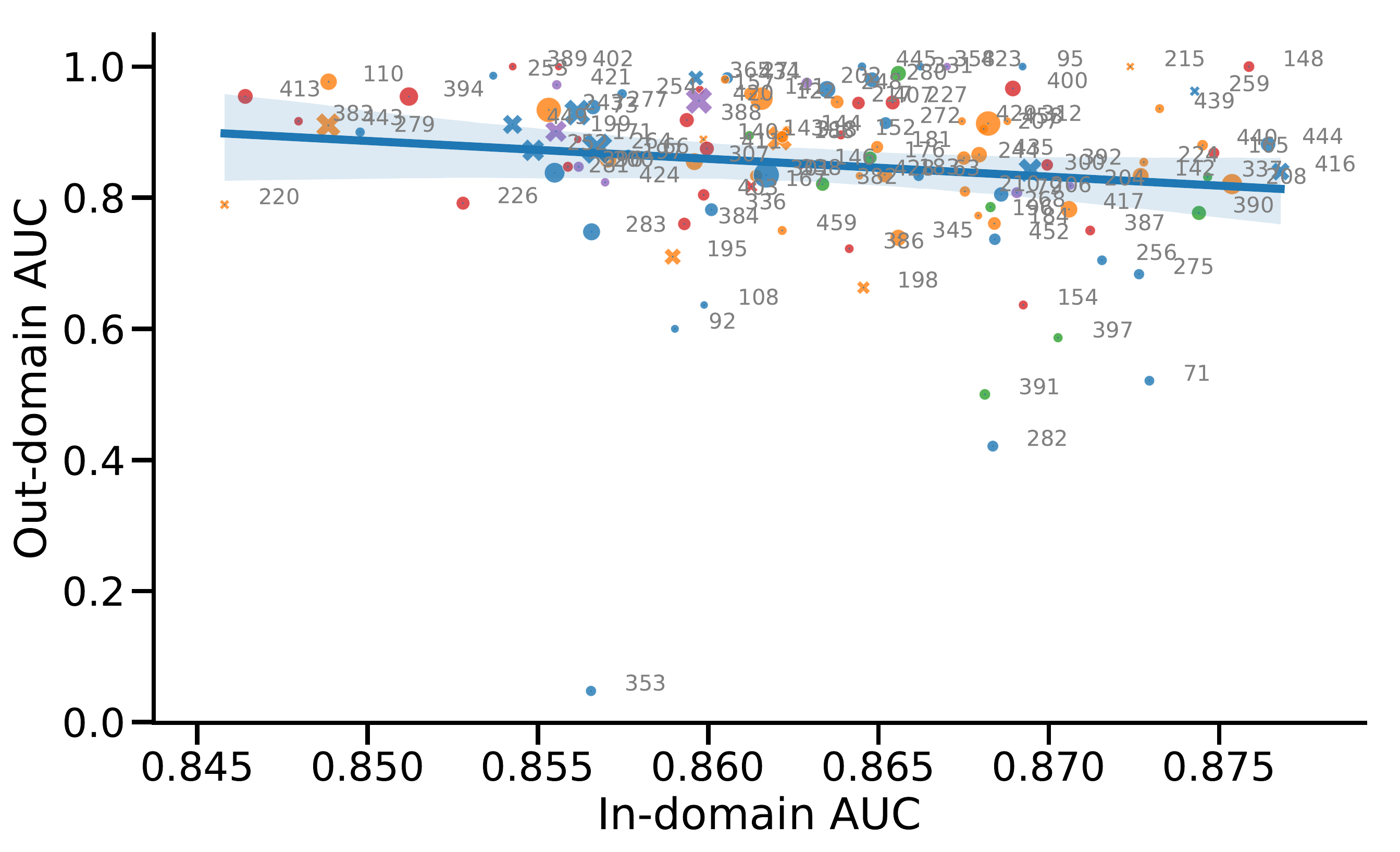}\hfill

\caption{\textbf{Correlations of OoD performance}.  Scatterplots comparing the Out-of-domain (test) AUC to gender imbalance, age, and In-domain AUC on the hospital level (text annotations denote hospital IDs). \texttt{Region\_group\_mode} corresponds to Region names in the US, while \texttt{region\_size} corresponds to the number of datapoints per hospitals. }
\label{fig:demographics_scatterplot}
\end{figure*}

Demographics play an important role in the performance of clinical models \citep{finlayson2021clinician} and therefore we investigate whether age or gender explain these differences. In figure \ref{fig:demographics_scatterplot}, we present scatterplots of each hospital as a function of average gender imbalance and age versus the test AUC. We fit linear regression models to illustrate these trends. We observe no trend between OoD test performance and gender imbalance. For age, we observe a very moderate negative trend between lower test AUC and average age: hospitals with older patients tend to perform worse out of domain.

In Figure \ref{fig:demographics_scatterplot}, we also plot the in-domain (test split of training set) versus the out-domain (test split of test set) performance on the hospital level. We observe a moderate negative trend between lower out-domain and higher in-domain AUC: hospitals perform worse OoD when the in-domain performance is higher, pointing to overfitting issues. In other words, these hospitals over-rely on the training data and struggle to generalize to new settings. We should note though, that the In-domain AUC has a very limited range (0.84-0.88 AUC) compared to the Out-domain AUC, which could confound this trend. This limited range is expected given that the in-domain test set is usually mostly the same, since we only leave out one hospital at a time - the train environment is usually almost the same.

\subsection{Cross-sectional features for OoD environment identification} \label{envs_demographics}

As an alternative to the model-based leave-one-hospital-out approach, we explored a method based on cross-sectional features. As discussed earlier, previous approaches focused on using single demographic features as exclusion criteria \citep{zadorozhny2021out}. However, by simply training on males and testing on females, for example, we will not be able to assess the ability of domain generalization algorithms to learn invariances across \emph{multiple} environments. To mitigate this, we take continuous and categorical features and calculate cross-sectional quantiles across the entire dataset. The hypothesis here is that specific cross-sections (e.g. young women) will show higher degrees of OoDness. 

First, we start with single features (age) and we split it into ten quantiles where each quantile has almost 3000 patients. By using the two oldest quantiles as test set, ERMID again underperforms ERM and ERMMerged, indicating that this environment is not OoD. Similar results are obtained with the youngest quantile. 

To create more complex and numerous environments, we then explored cross-sectional feature environments by intersecting continuous and categorical features (e.g. age and gender) to produce equally-sized quantiles. We split the patient stays into twenty cross-sectional age and gender quantiles, where each quantile has almost 1500 patients. Following a similar resampling procedure as in the previous section, we found that when using `young women' as test set, the ERMID model slightly overperforms the rest models (0.82 [0.78-0.86], over 0.81 [0.77-0.85] for ERM, and 0.81 [0.77-0.86]), which, however given the CI overlaps must be considered marginal. The `old women' environment did not show any OoD properties.

\newpage
\subsection{Supplementary figures }

\input{table_splits}

\begin{figure*}
    \centering
    \includegraphics[width=0.8\textwidth]{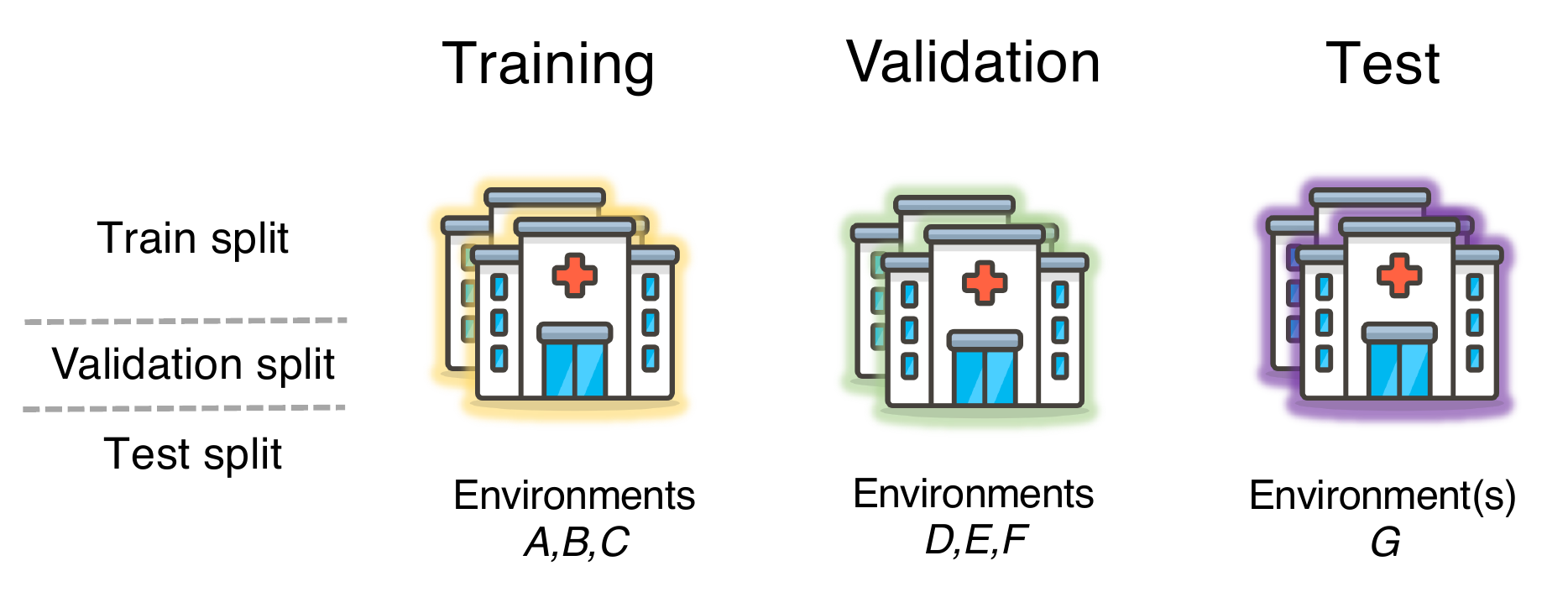}
   
    \caption{\textbf{Illustration of the environment selection strategy}. Each environment (e.g. hospital) is assigned to a single \textit{vertical } set. Within each set there are further \textit{horizontal} splits to enable in-domain and out-of-domain comparisons.}
    \label{fig:splits}
    
\end{figure*}

\let\clearpage\relax

\end{document}

%% file: table_resampling.tex




\begin{table}[]
\resizebox{0.5\textwidth}{!}{
\begin{tabular}{@{}lcc@{}}
\toprule
\textbf{Model} & \textbf{Imbalanced}  & \textbf{Resampled}   \\ \midrule
ERM            & 0.71 {[}0.63-0.79{]} & 0.68 {[}0.60-0.76{]} \\
ERMMerged      & 0.72 {[}0.64-0.79{]} & 0.69 {[}0.61-0.77{]} \\
ERMID          & 0.71 {[}0.64-0.78{]} & 0.67 {[}0.59-0.74{]} \\ \bottomrule
\end{tabular}
}
\caption{\textbf{Impact of resampling}. Comparing the training scenarios as a function of access to \textit{imbalanced} versus \textit{subsampled} training sets that match ERMID (mean AUC and 95\% CIs).}
\label{tab:resampling}
 \vspace{-0.7cm}
\end{table}

%% file: table_parameterization.tex
\begin{table}[]
\resizebox{0.5\textwidth}{!}{
\begin{tabular}{@{}lcc@{}}
\toprule
\multirow{2}{*}{\textbf{Model}} & \multicolumn{2}{c}{\textbf{Resampled}}                                                           \\ \cmidrule(l){2-3} 
                                & \multicolumn{1}{l}{\textbf{Underparameterized}} & \multicolumn{1}{l}{\textbf{Overparameterized}} \\ \midrule
ERM                             & 0.73 {[}0.66-0.80{]}                            & 0.68 {[}0.60-0.76{]}                           \\
ERMMerged                       & 0.72 {[}0.64-0.79{]}                            & 0.69 {[}0.61-0.77{]}                           \\
ERMID                           & 0.70 {[}0.63-0.78{]}                            & 0.67 {[}0.59-0.74{]}                           \\ \bottomrule
\end{tabular}
}
\caption{\textbf{Impact of large models}. Comparing the training scenarios as a function of using \textit{underparameterized} (small) versus \textit{overparameterized} (large) models. Both variants use the resampled setup of Table \ref{tab:resampling} and hence the right columns of both tables are identical (mean AUC and 95\% CIs).}
\label{tab:param}
 \vspace{-0.7cm}
\end{table}

%% file: table_splits.tex
\begin{table*}
\centering
\resizebox{0.8\textwidth}{!}{

\begin{tabular}{@{}lcccll@{}}
\toprule
\multirow{2}{*}{\textbf{Split}} & \multirow{2}{*}{\textbf{\begin{tabular}[c]{@{}c@{}}Hospital \\ size\end{tabular}}} & \multirow{2}{*}{\textbf{Age}} & \multirow{2}{*}{\textbf{\begin{tabular}[c]{@{}c@{}}Gender \\ balance\end{tabular}}} & \multicolumn{2}{c}{\textbf{Regions}}                                                                                                                                  \\ \cmidrule(l){5-6} 
                                &                                                                                    &                               &                                                                                     & \multicolumn{1}{c}{\textbf{Name}}                                                                      & \multicolumn{1}{c}{\textbf{Count}}                           \\ \midrule
Train                           & 261.07 ± 244.06                                                                    & 64.17 ± 3.03                  & 0.54  ± 0.05                                                                        & \begin{tabular}[c]{@{}l@{}}South        \\ Midwest \\ West        \\ Northeast \\ Missing\end{tabular} & \begin{tabular}[c]{@{}l@{}}31\\ 26\\ 21\\ 9\\ 6\end{tabular} \\ \cmidrule(l){5-6} 
Validation                      & 153.62 ± 99.22                                                                     & 66.01 ± 1.38                  & 0.54 ± 0.08                                                                         & \begin{tabular}[c]{@{}l@{}}Midwest    \\ Missing    \\ South      \\ West\end{tabular}                 & \begin{tabular}[c]{@{}l@{}}3\\ 2\\ 2\\ 1\end{tabular}        \\ \cmidrule(l){5-6} 
Test                            & 180.91 ± 114.43                                                                    & 65.27 ± 4.01                  & 0.55 ± 0.04                                                                         & \begin{tabular}[c]{@{}l@{}}South      \\ Midwest    \\ West\end{tabular}                               & \begin{tabular}[c]{@{}l@{}}5\\ 4\\ 3\end{tabular}            \\ \bottomrule
\end{tabular}
}
\caption{\textbf{Characteristics of each split aggregated by hospital after applying Algorithm \ref{algorithm}.} Hospital size and age correspond to the mean ($\pm$ std) patient number and age per hospital, respectively. Gender balance is the mean ($\pm$ std) male prevalence (0.5 is balanced). The last column shows the number of hospitals from each region. }
\label{tab:sets_statistics}
\end{table*}

%% file: sample-authordraft.bbl
\begin{thebibliography}{24}
\providecommand{\natexlab}[1]{#1}
\providecommand{\url}[1]{\texttt{#1}}
\expandafter\ifx\csname urlstyle\endcsname\relax
  \providecommand{\doi}[1]{doi: #1}\else
  \providecommand{\doi}{doi: \begingroup \urlstyle{rm}\Url}\fi

\bibitem[Arjovsky et~al.(2019)Arjovsky, Bottou, Gulrajani, and
  Lopez-Paz]{arjovsky2019invariant}
Martin Arjovsky, L{\'e}on Bottou, Ishaan Gulrajani, and David Lopez-Paz.
\newblock Invariant risk minimization.
\newblock \emph{arXiv preprint arXiv:1907.02893}, 2019.

\bibitem[Beery et~al.(2018)Beery, Van~Horn, and Perona]{beery2018recognition}
Sara Beery, Grant Van~Horn, and Pietro Perona.
\newblock Recognition in terra incognita.
\newblock In \emph{Proceedings of the European conference on computer vision
  (ECCV)}, pages 456--473, 2018.

\bibitem[Castro et~al.(2020)Castro, Walker, and Glocker]{castro2020causality}
Daniel~C Castro, Ian Walker, and Ben Glocker.
\newblock Causality matters in medical imaging.
\newblock \emph{Nature Communications}, 11\penalty0 (1):\penalty0 1--10, 2020.

\bibitem[Chung et~al.(2014)Chung, Gulcehre, Cho, and
  Bengio]{chung2014empirical}
Junyoung Chung, Caglar Gulcehre, KyungHyun Cho, and Yoshua Bengio.
\newblock Empirical evaluation of gated recurrent neural networks on sequence
  modeling.
\newblock \emph{arXiv preprint arXiv:1412.3555}, 2014.

\bibitem[Dai and Van~Gool(2018)]{dai2018dark}
Dengxin Dai and Luc Van~Gool.
\newblock Dark model adaptation: Semantic image segmentation from daytime to
  nighttime.
\newblock In \emph{2018 21st International Conference on Intelligent
  Transportation Systems (ITSC)}, pages 3819--3824. IEEE, 2018.

\bibitem[Desautels et~al.(2017)Desautels, Calvert, Hoffman, Mao, Jay, Fletcher,
  Barton, Chettipally, Kerem, and Das]{desautels2017using}
Thomas Desautels, Jacob Calvert, Jana Hoffman, Qingqing Mao, Melissa Jay, Grant
  Fletcher, Chris Barton, Uli Chettipally, Yaniv Kerem, and Ritankar Das.
\newblock Using transfer learning for improved mortality prediction in a
  data-scarce hospital setting.
\newblock \emph{Biomedical informatics insights}, 9:\penalty0 1178222617712994,
  2017.

\bibitem[Finlayson et~al.(2021)Finlayson, Subbaswamy, Singh, Bowers, Kupke,
  Zittrain, Kohane, and Saria]{finlayson2021clinician}
Samuel~G Finlayson, Adarsh Subbaswamy, Karandeep Singh, John Bowers, Annabel
  Kupke, Jonathan Zittrain, Isaac~S Kohane, and Suchi Saria.
\newblock The clinician and dataset shift in artificial intelligence.
\newblock \emph{The New England journal of medicine}, 385\penalty0
  (3):\penalty0 283, 2021.

\bibitem[Gulrajani and Lopez-Paz(2020)]{gulrajani2020search}
Ishaan Gulrajani and David Lopez-Paz.
\newblock In search of lost domain generalization.
\newblock \emph{arXiv preprint arXiv:2007.01434}, 2020.

\bibitem[Guo et~al.(2022)Guo, Pfohl, Fries, Johnson, Posada, Aftandilian, Shah,
  and Sung]{guo2022evaluation}
Lin~Lawrence Guo, Stephen~R Pfohl, Jason Fries, Alistair~EW Johnson, Jose
  Posada, Catherine Aftandilian, Nigam Shah, and Lillian Sung.
\newblock Evaluation of domain generalization and adaptation on improving model
  robustness to temporal dataset shift in clinical medicine.
\newblock \emph{Scientific reports}, 12\penalty0 (1):\penalty0 1--10, 2022.

\bibitem[Hyland et~al.(2020)Hyland, Faltys, H{\"u}ser, Lyu, Gumbsch, Esteban,
  Bock, Horn, Moor, Rieck, et~al.]{hyland2020early}
Stephanie~L Hyland, Martin Faltys, Matthias H{\"u}ser, Xinrui Lyu, Thomas
  Gumbsch, Crist{\'o}bal Esteban, Christian Bock, Max Horn, Michael Moor,
  Bastian Rieck, et~al.
\newblock Early prediction of circulatory failure in the intensive care unit
  using machine learning.
\newblock \emph{Nature medicine}, 26\penalty0 (3):\penalty0 364--373, 2020.

\bibitem[Johnson et~al.(2016)Johnson, Pollard, Shen, Lehman, Feng, Ghassemi,
  Moody, Szolovits, Anthony~Celi, and Mark]{johnson2016mimic}
Alistair~EW Johnson, Tom~J Pollard, Lu~Shen, Li-wei~H Lehman, Mengling Feng,
  Mohammad Ghassemi, Benjamin Moody, Peter Szolovits, Leo Anthony~Celi, and
  Roger~G Mark.
\newblock Mimic-iii, a freely accessible critical care database.
\newblock \emph{Scientific data}, 3\penalty0 (1):\penalty0 1--9, 2016.

\bibitem[Johnson et~al.(2018)Johnson, Pollard, and
  Naumann]{johnson2018generalizability}
Alistair~EW Johnson, Tom~J Pollard, and Tristan Naumann.
\newblock Generalizability of predictive models for intensive care unit
  patients.
\newblock \emph{arXiv preprint arXiv:1812.02275}, 2018.

\bibitem[Koh et~al.(2021)Koh, Sagawa, Marklund, Xie, Zhang, Balsubramani, Hu,
  Yasunaga, Phillips, Gao, et~al.]{koh2021wilds}
Pang~Wei Koh, Shiori Sagawa, Henrik Marklund, Sang~Michael Xie, Marvin Zhang,
  Akshay Balsubramani, Weihua Hu, Michihiro Yasunaga, Richard~Lanas Phillips,
  Irena Gao, et~al.
\newblock Wilds: A benchmark of in-the-wild distribution shifts.
\newblock In \emph{International Conference on Machine Learning}, pages
  5637--5664. PMLR, 2021.

\bibitem[Li et~al.(2018)Li, Yang, Song, and Hospedales]{li2018learning}
Da~Li, Yongxin Yang, Yi-Zhe Song, and Timothy~M Hospedales.
\newblock Learning to generalize: Meta-learning for domain generalization.
\newblock In \emph{Thirty-Second AAAI Conference on Artificial Intelligence},
  2018.

\bibitem[McDermott et~al.(2021)McDermott, Nestor, Kim, Zhang, Goldenberg,
  Szolovits, and Ghassemi]{mcdermott2021comprehensive}
Matthew McDermott, Bret Nestor, Evan Kim, Wancong Zhang, Anna Goldenberg, Peter
  Szolovits, and Marzyeh Ghassemi.
\newblock A comprehensive ehr timeseries pre-training benchmark.
\newblock In \emph{Proceedings of the Conference on Health, Inference, and
  Learning}, pages 257--278, 2021.

\bibitem[Nestor et~al.(2019)Nestor, McDermott, Boag, Berner, Naumann, Hughes,
  Goldenberg, and Ghassemi]{nestor2019feature}
Bret Nestor, Matthew~BA McDermott, Willie Boag, Gabriela Berner, Tristan
  Naumann, Michael~C Hughes, Anna Goldenberg, and Marzyeh Ghassemi.
\newblock Feature robustness in non-stationary health records: caveats to
  deployable model performance in common clinical machine learning tasks.
\newblock In \emph{Machine Learning for Healthcare Conference}, pages 381--405.
  PMLR, 2019.

\bibitem[Pollard et~al.(2018)Pollard, Johnson, Raffa, Celi, Mark, and
  Badawi]{pollard2018eicu}
Tom~J Pollard, Alistair~EW Johnson, Jesse~D Raffa, Leo~A Celi, Roger~G Mark,
  and Omar Badawi.
\newblock The eicu collaborative research database, a freely available
  multi-center database for critical care research.
\newblock \emph{Scientific data}, 5\penalty0 (1):\penalty0 1--13, 2018.

\bibitem[Rocheteau et~al.(2021)Rocheteau, Li{\`o}, and
  Hyland]{rocheteau2021temporal}
Emma Rocheteau, Pietro Li{\`o}, and Stephanie Hyland.
\newblock Temporal pointwise convolutional networks for length of stay
  prediction in the intensive care unit.
\newblock In \emph{Proceedings of the Conference on Health, Inference, and
  Learning}, pages 58--68, 2021.

\bibitem[Sheikhalishahi et~al.(2020)Sheikhalishahi, Balaraman, and
  Osmani]{sheikhalishahi2020benchmarking}
Seyedmostafa Sheikhalishahi, Vevake Balaraman, and Venet Osmani.
\newblock Benchmarking machine learning models on multi-centre eicu critical
  care dataset.
\newblock \emph{Plos one}, 15\penalty0 (7):\penalty0 e0235424, 2020.

\bibitem[Singh et~al.(2022)Singh, Mhasawade, and
  Chunara]{singh2022generalizability}
Harvineet Singh, Vishwali Mhasawade, and Rumi Chunara.
\newblock Generalizability challenges of mortality risk prediction models: A
  retrospective analysis on a multi-center database.
\newblock \emph{PLOS Digital Health}, 1\penalty0 (4):\penalty0 e0000023, 2022.

\bibitem[Sjoding et~al.(2020)Sjoding, Dickson, Iwashyna, Gay, and
  Valley]{sjoding2020racial}
Michael~W Sjoding, Robert~P Dickson, Theodore~J Iwashyna, Steven~E Gay, and
  Thomas~S Valley.
\newblock Racial bias in pulse oximetry measurement.
\newblock \emph{New England Journal of Medicine}, 383\penalty0 (25):\penalty0
  2477--2478, 2020.

\bibitem[Wiens et~al.(2014)Wiens, Guttag, and Horvitz]{wiens2014study}
Jenna Wiens, John Guttag, and Eric Horvitz.
\newblock A study in transfer learning: leveraging data from multiple hospitals
  to enhance hospital-specific predictions.
\newblock \emph{Journal of the American Medical Informatics Association},
  21\penalty0 (4):\penalty0 699--706, 2014.

\bibitem[Zadorozhny et~al.(2021)Zadorozhny, Thoral, Elbers, and
  Cin{\`a}]{zadorozhny2021out}
Karina Zadorozhny, Patrick Thoral, Paul Elbers, and Giovanni Cin{\`a}.
\newblock Out-of-distribution detection for medical applications: Guidelines
  for practical evaluation.
\newblock \emph{arXiv preprint arXiv:2109.14885}, 2021.

\bibitem[Zhang et~al.(2021)Zhang, Dullerud, Seyyed-Kalantari, Morris, Joshi,
  and Ghassemi]{zhang2021empirical}
Haoran Zhang, Natalie Dullerud, Laleh Seyyed-Kalantari, Quaid Morris, Shalmali
  Joshi, and Marzyeh Ghassemi.
\newblock An empirical framework for domain generalization in clinical
  settings.
\newblock In \emph{Proceedings of the Conference on Health, Inference, and
  Learning}, pages 279--290, 2021.

\end{thebibliography}
